\documentclass[letterpaper, 10 pt, conference]{ieeeconf}  

\IEEEoverridecommandlockouts                              

\usepackage{epsfig}
\usepackage{graphicx}
\usepackage{amsmath}
\usepackage{amssymb}
\usepackage{graphicx} 
\usepackage{times} 
\usepackage{bm}
\usepackage{subfig}
\usepackage{url}
\usepackage{multirow}
\usepackage{float}
\usepackage{textcomp}

\DeclareMathOperator*{\argmin}{argmin}

\begin{document}

\title{Combining Deep Learning with Geometric Features for Image based Localization in the Gastrointestinal Tract }

\author{
Jingwei Song, Mitesh Patel, Andreas Girgensohn, Chelhwon Kim \\
FX Palo Alto Laboratory Inc. \\
Palo Alto, CA - 94304, USA\\
{\tt\small jingweisong.eng@gmail.com}\\
{\tt\small {mitesh, andreasg, kim}@fxpal.com}
}

\maketitle

\begin{abstract}

Tracking monocular colonoscope in the Gastrointestinal tract (GI) is a challenging problem as the images suffer from deformation, blurred textures, significant changes in appearance. They greatly restrict the tracking ability of conventional geometry based methods. Even though Deep Learning (DL) can overcome these issues, limited labeling data is a roadblock to state-of-art DL method. Considering these, we propose a novel approach to combine DL method with traditional feature based approach to achieve better localization with small training data. Our method fully exploits the best of both worlds by introducing a Siamese network structure to perform few-shot classification to the closest zone in the segmented training image set. The classified label is further adopted to initialize the pose of scope. To fully use the training dataset, a pre-generated triangulated map points within the zone in the training set are registered with observation and contribute to estimating the optimal pose of the test image. The proposed hybrid method is extensively tested and compared with existing methods, and the result shows significant improvement over traditional geometric based or DL based localization. The accuracy is improved by $28.94\%$ (Position) and $10.97\%$ (Orientation) with respect to state-of-art method.

\end{abstract}

\section{Introduction}

Endoscopic systems offer a minimally invasive way to examine the internal body structures and hence their applications are rapidly extending to cover the increasing needs for accurate therapeutic interventions~\cite{sganga2018offsetnet, urban:2018, iakovidis:2018}. Due to the non-invasive way of screening, such systems allows medical practitioners to observe early stage precancerous polyps during the screening process and/or cancerous lung nodules in case of bronchoscopy~\cite{sganga2018offsetnet}. 
It is of paramount importance for such systems to accurately localize and track itself in a given GI tract and/or bronchi tract to provide the location of various pathological findings to the medical practitioners.\par

\begin{figure}
        \centering 
        \includegraphics[width=0.49\textwidth]{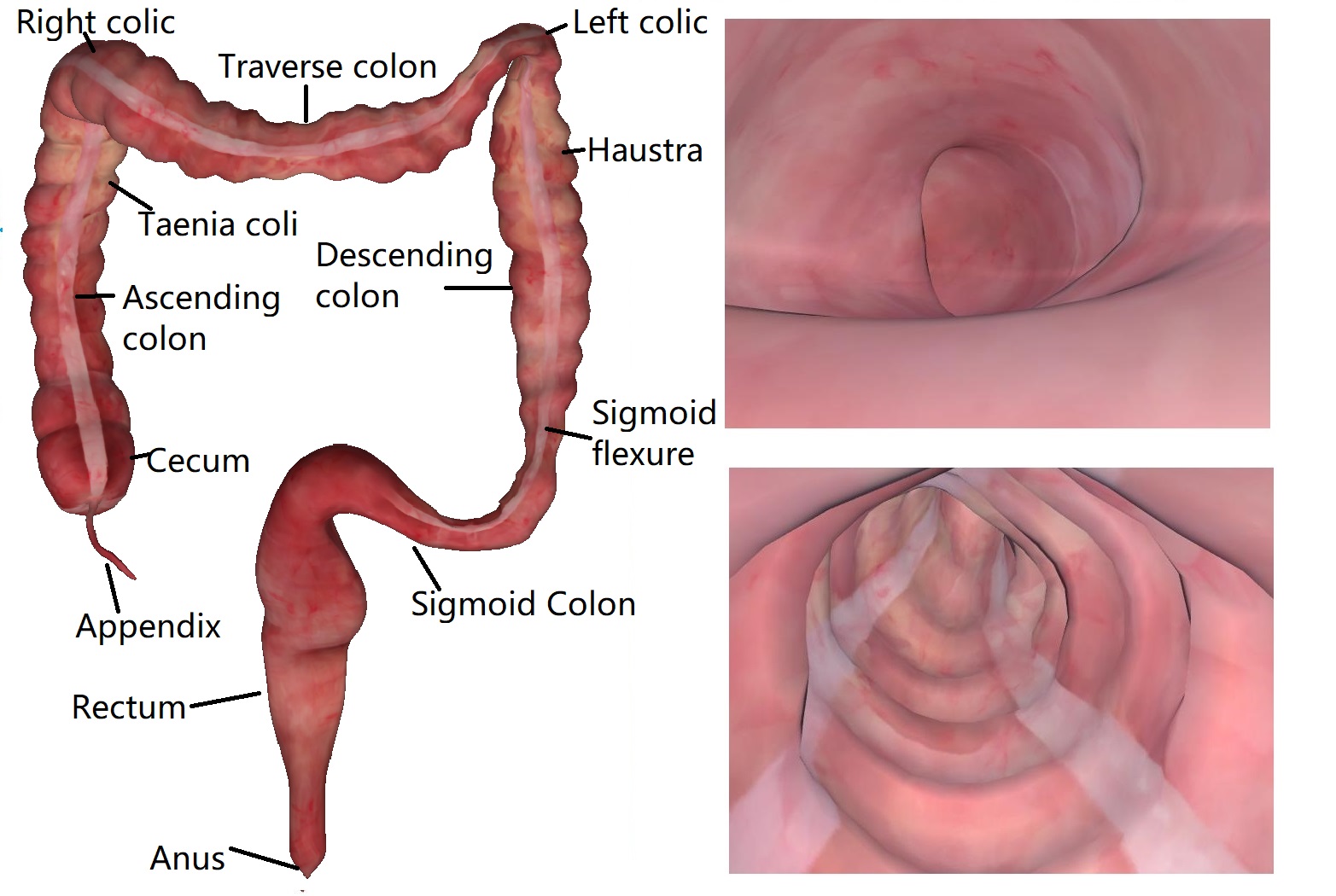}
        \caption{The left is the 3D ex-vivo phantom model used for data generation. The right images are sample data yield by the simulator. } 
        \label{fig:simulationdateaset_gen}
\end{figure}

To date, several researchers have explored hardware level sensing modalities that can assist in localization, including Optitrack \cite{point2011optitrack}, programmable robot arm \cite{iakovidis:2018} or Electro-magnetic sensor \cite{wang2019novel}. However, due to the difficulties in multiple sensor fusion or high cost, image-based localization is still popular in CAS and/or screening applications for medical practitioners. Various techniques like image to CT registration \cite{ingram2017feasibility}, Simultaneous Localization And Mapping (SLAM) \cite{mahmoud2016orbslam}, Visual Odometry (VO) \cite{turan2018deepvo} and deep learning (DL)  \cite{iakovidis:2018} are analyzed. Moreover, image-based localization has gained popularity due to the progress of localization techniques in the computer vision community and limited access to hardware.\par  

The state-of-art approaches include feature based~\cite{luo2012development,luo2014discriminative,shen2015robust,mahmoud2016orbslam,dimas2017scale}, DL based methods~\cite{sganga2018offsetnet,turan2018deepvo} and fusing DL with geometric \cite{turan2018deepvo}. The feature based methods apply various types of corner points tracked within images, and optimize pose with strict geometric information. The performance of these approaches is highly dependent on the quality of the registered corner points. Thus they are advantageous in texture rich environment instead of textureless scenario like GI tract. On the contrary, DL based methods, does not perform feature tracking process but model the feature extraction and geometry based optimization with a Convolutional Neural Network (CNN). These methods are advantageous over conventional geometric techniques as they are capable of exploiting more information. To overcome overfitting, large dataset is essential which is insufficient for medical domain due to privacy, cost or difficulty in data collection. Very limited works can be found in fusing DL with geometric method in CAS. However, it has be explored and proved efficient method in other fields such as indoor localization \cite{taira:2018inloc} and autonomous driving \cite{li2019pose}.\par

In this paper, we propose a novel framework that fuses DL technique with traditional feature based method to get the best of both worlds. The intuition is to initialize pose with DL based classification and refine with the geometric constraint. The hybrid system can be achieved in few-shot learning manner. Our system is capable to provide better pose estimates comparing with baseline methods.

\section{Related Work}

GI tract is heavily investigated by researchers in medical, computer vision and robotic communities~\cite{iakovidis:2018, sganga2018offsetnet,song20163d,del2009combined}. The techniques can be categorized into three groups, i.e. geometric feature based, DL based and hybrid of DL and geometric techniques.

\textbf{Geometric based method} is a technique that extracts and registers key corner points from images; the relative poses of these images are estimated by enforcing the geometric constraints over the tracked key points. \cite{luo2012development} develops a hybrid algorithm for predicting the motion of bronchoscope using epipolar constraint, and Kalman filter to estimate the magnitude of the motion. \cite{luo2014discriminative} improves it by a discriminative structural similarity measure to boost video-volume registration which leads to better results. Later, \cite{mahmoud2016orbslam} proves that ORB-SLAM \cite{mur2017orb}, an off-the-shelf SLAM framework, that can estimate both the scope location and the 3D structure of the scene. Different from 2D to 2D image matching technique, \cite{shen2015robust} adopts Shape from Shading (SfS) to directly convert a single scope image to 3D space and register the shape with 3D image. They claim that the structural information in the 2D image is fully exploited.\par

\textbf{DL} has been widely applied in pose regression~\cite{kendall2015posenet, weyand2016planet, arandjelovic2016netvlad, radwan2018vlocnet}. CNN or its variants are used to learn features from the image content and is optimized towards a better representation that fulfills specific tasks. DL is adopted to localize scopes within GI tract~\cite{iakovidis:2018, turan2018deepvo} and bronchial tract~\cite{sganga2018offsetnet} which outperforms feature based techniques. \cite{turan2018deepvo} proposes a monocular VO method which uses a recurrent CNN for feature extraction. \cite{dimas2017scale} uses a DL framework to estimate the scale to get better pose accuracy. Lastly, \cite{sganga2018offsetnet} proposes a system called \emph{Offsetnet} which performs image-based localization in bronchial tract. Despite its success, DL requires large dataset for training the models which is hard to obtain in medical domain. As \cite{szegedy2015going} points out, high dimension of parameters makes the enlarged network prone to overfitting. The base networks involved in \cite{turan2018deepvo, iakovidis:2018,sganga2018offsetnet} are all multiple layer deep networks.\par

\textbf{Hybrid of DL and traditional geometric methods} has also been explored for image-based localization~\cite{taira:2018inloc}\cite{li2019pose}. \cite{taira:2018inloc} proposes a pose estimation technique that fuses DL with feature based methods for indoor localization. It locates the image to a given area/zone, and estimates the pose with geometric approach. Moreover, they utilize depth information to estimate the pose. In CAS, very limited work is introduced. The work of \cite{turan2018deepvo} can be reluctantly categorized in this field because it employs a structure from motion neural network named SFM-learner \cite{zhou2017unsupervised} to consider geometric relation. Note that this SFM-learner is still a DL method and requires descent amount of training data.\par 


In this paper, we overcome the drawbacks, overfitting in DL and losing tracking in geometric methods, by developing a hybrid system of DL and geometric methods. We are not aware of any similar approach in CAS. To the best of our knowledge the system proposed by \cite{sganga2018offsetnet} is the closest to ours in CAS, while the main difference is that they utilize DL to bridge the pose of Computer Tomography (CT) and real image but without considering the real geometric relations. The novelties of our proposed system are summarized below:

\begin{itemize}
    \item We propose a hybrid system that fuses the DL method with traditional geometric methods. Our hybrid approach requires small dataset for training and achieves optimal results.  
    \item We employ Siamese network allowing few-shot learning. Our system is modular to incorporate expert knowledge from medical practitioner within our system.
    \item Our hybrid system is robust to noisy images. We provide a way to filter outliers and reject bad results.
\end{itemize}

\begin{figure*}[]
        \centering
        \includegraphics[width=0.95\textwidth]{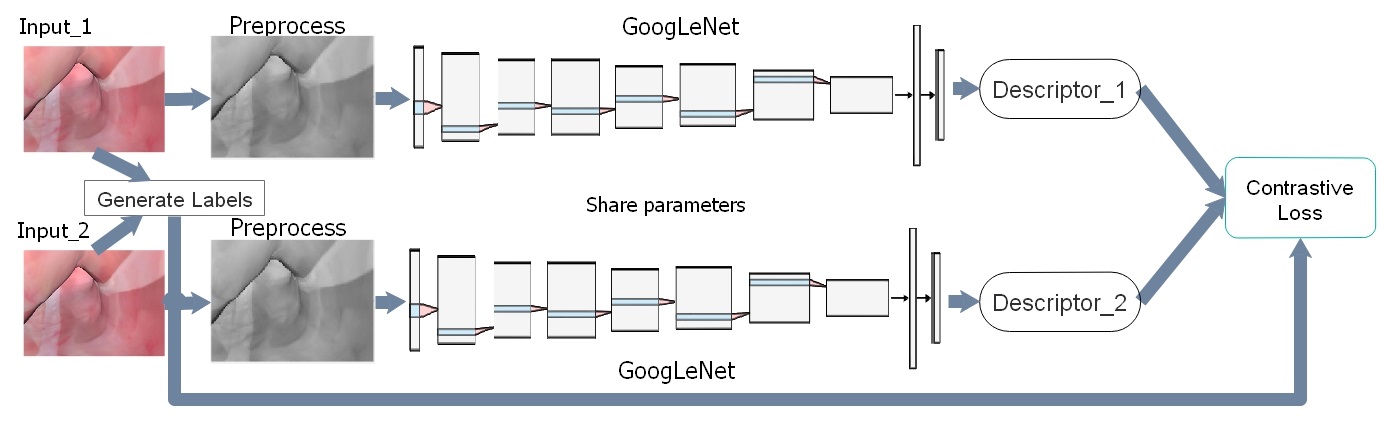}
        \caption{The structure of Siamese network for image zone classification.}
        \label{fig:Siamese_network}
    \end{figure*}
\begin{figure*}[htpb]    
    \centering
    \subfloat{    
    }\\
    \subfloat{    
        \begin{minipage}[htpb]{1\textwidth}    
            \centering
            \includegraphics[width=1\linewidth]{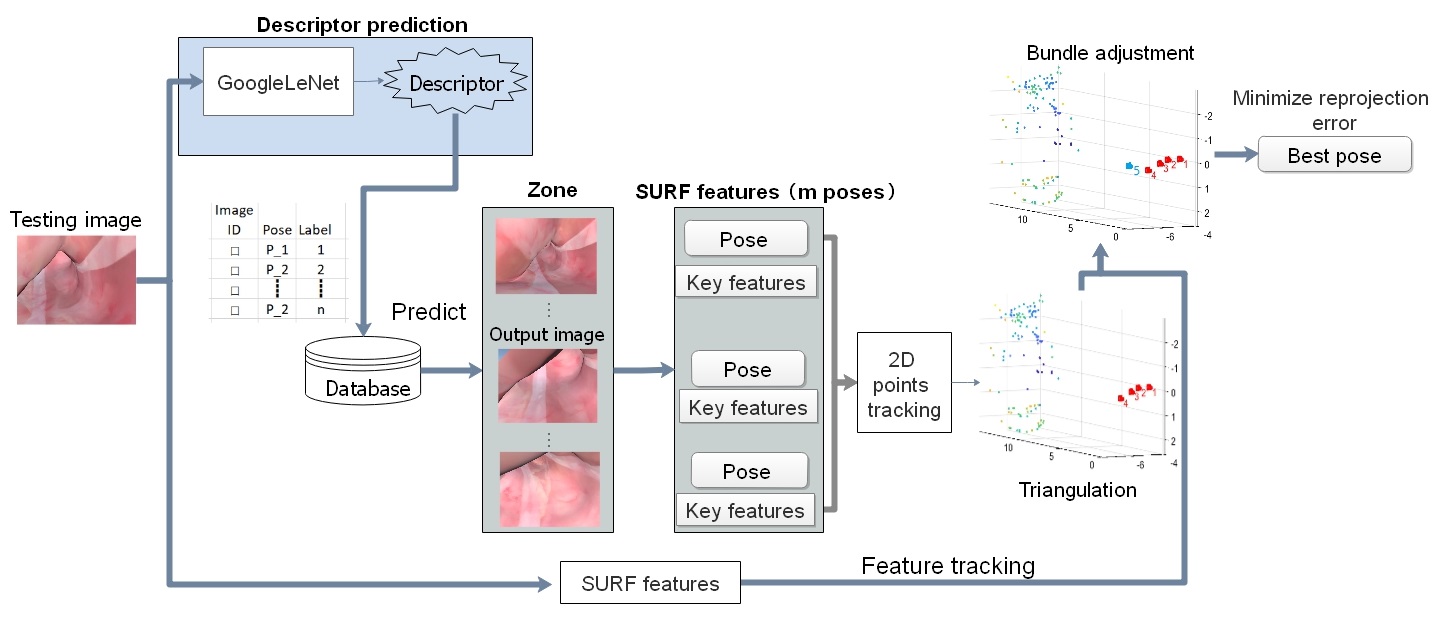}            
        \end{minipage}                
    }
    \caption{Illustrated is the proposed framework. The block in blue is DL based image classification, and the rest shows the two-step pose refinement with geometric constraints.}
    \label{fig:Flowchart}
\end{figure*}

\section{Proposed Methodology}

The framework consists two steps: (a) zone classification and (b) two-step pose refinement. The zone classification is a DL framework that predicts the zone to which a test image is captured. For this, we utilize a Siamese network based DL pipeline to learn the image similarity between the test image and the zone. The pose of the test image is initialized with the median of the zone. In the two-step pose refinement step, the classified zone information is further utilized to estimate the actual pose of the test image. Images and related poses within the zone are triangulated to generate map points, and the pose of the test image is further refined by going through a two-step pose refinement. Specifically, we use the images from the zone to perform multiple image triangulation to yield map points in local coordinate frame, and map points are further registered to the test image. The pose of the test image is optimized by minimizing the re-projection errors between the registered 3D map points and 2D image pixels. An overview of our proposed system can be found in Figure~\ref{fig:Flowchart}. \par 

We would like to address that our localization framework is based on 'learning and inferencing' workflow where the global pose is required in the training step. Unlike researches \cite{shen2015robust, luo2014discriminative, sganga2018offsetnet, luo2012development} bridging the prior 3D image (CT scan) with 2D image-based observation, our work aims at image-based localization in the scenario with global pose retrieved from some sources like Optitrack \cite{turan2018deepvo}, robot arm \cite{iakovidis:2018} or Electro-magnetic sensor \cite{wang2019novel}. In the case of CT scan as the training data in \cite{shen2015robust, luo2014discriminative, sganga2018offsetnet, luo2012development}, the image poses labels can be retrieved offline with similarity based automatic image to CT registration \cite{luo2012development,luo2014discriminative,ingram2017feasibility}, or from semi-automatic pose labeling, because our work only requires small training dataset.\par

\subsection{Classification using DL}
\label{section_classification_Siamese}

For zone classification, we use a Siamese neural network which learns the image descriptors. Siamese architecture is first introduced by~\cite{bromley1994signature} and used for signature verification by matching images. Since then, it has widely been applied for various image similarity problems such as semantic similarity and object tracking. A Siamese network consists a twin neural networks which share the same architecture and same set of weights, and hence the distance between the descriptors of similar images is smaller compare with dis-similar images. Further, due to the characteristics of comparing images, the Siamese network can be trained with a smaller dataset. To balance the training data, the positive pairs should be equal to negative pairs. Therefore, for $m$ images and $n$ zones, the uppermost positive images pairs to be generated is $n\binom{m}{2}$ samples ($\binom{i}{j}$ is the binomial coefficient) much more than $\mathrm{m}\mathrm{n}$ samples for DL training.\par 

The Siamese neural network used in our experiments is shown in Figure~\ref{fig:Siamese_network}. PoseNet~\cite{kendall2015posenet} is adopted as the base network which is a GoogLeNet~\cite{szegedy2015going} architecture. The structure consists 23 CNN layers and one fully-connected layer which generates an output vector of size 128. As shown in \ref{fig:Siamese_network}, two same networks shares the weights. The images used in our system are normalized and resized to $224\time 224 \time 1 $. Each twin network takes one of the two binary images and applies two GoogLeNet with ReLU activation function and max-pooling layers. Branch outputs are concatenated and fed into linearly fully connected layers to generate the feature vectors denoted as descriptors. We use a contrastive loss function \cite{hadsell2006dimensionality} in Eq. (\ref{Eq_constrastive_loss}) to measure the similarity between the encoded descriptors. In Eq. (\ref{Eq_constrastive_loss}), $\mathbf{x}$, $\Tilde{\mathbf{x}}$ and $\mathrm{m}$ is the ground truth, prediction and the prior margin respectively. During inference, the test image is looked up in the training image descriptor database to find the most similar zone with the constrastive loss. The zone index with smallest loss is assigned as the zone to the test image.
\begin{equation}
\label{Eq_constrastive_loss}
\begin{aligned} 
&L\left(\mathbf{x}, \Tilde{\mathbf{x}}\right)=\\
&(1-\mathbf{x}) \frac{1}{2}\left(\Tilde{\mathbf{x}}\right)^{2}+(\mathbf{x}) \frac{1}{2}\left\{\max \left(0, \mathrm{m}-\Tilde{\mathbf{x}}\right)\right\}^{2} \end{aligned}
\end{equation}

\subsection{Two-step Pose Refinement}

\begin{figure}[]
        \centering
        \includegraphics[width=0.47\textwidth]{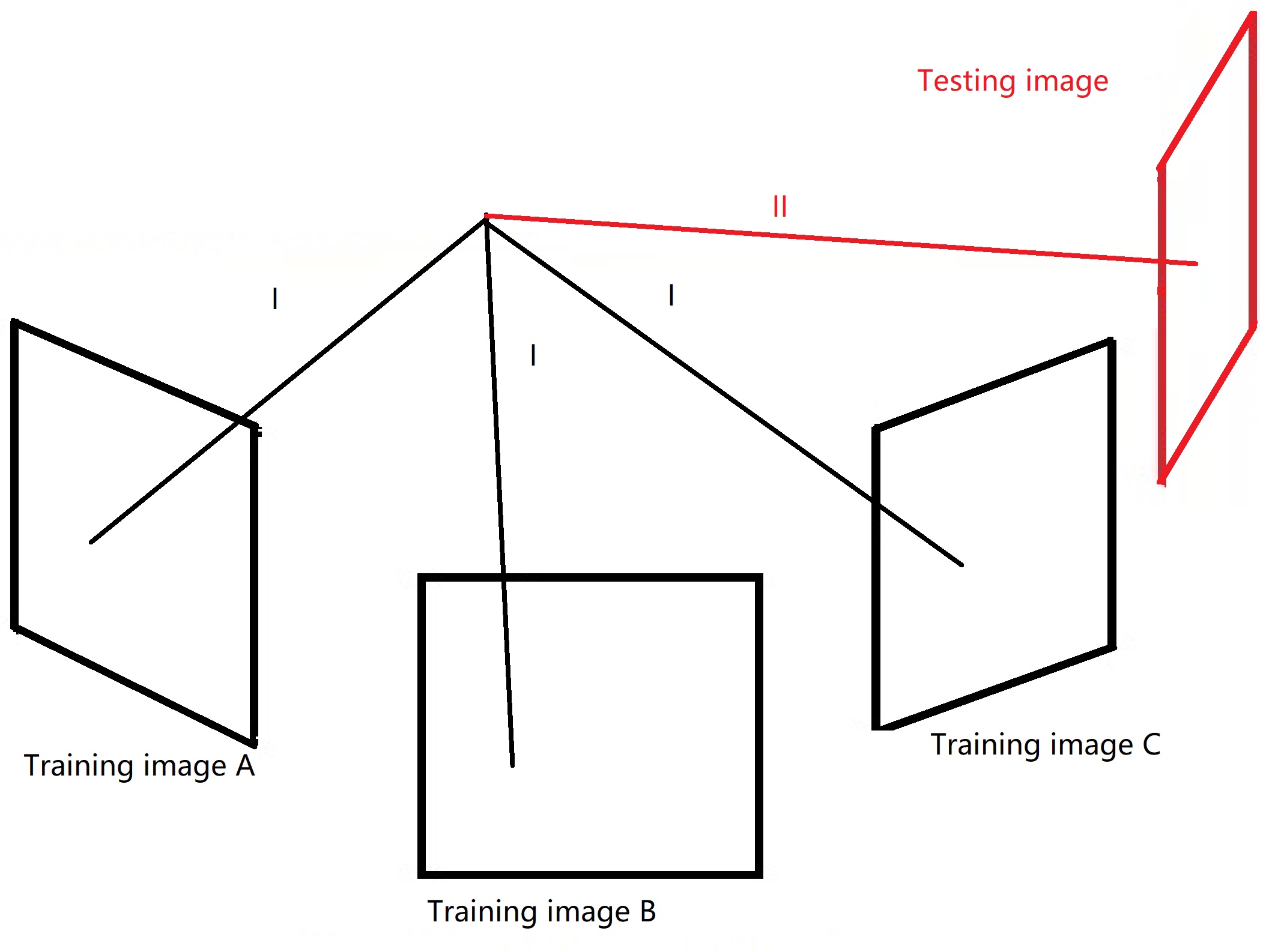}
        \caption{A diagram showing the two-step pose refinement process. In step I (in black), the training images (A-C) within one zone are used to track key 2D points. The tracked points and the known poses from the training images are triangulated to generate 3D map points. In step II (in red), after the test image is assigned to this zone, the 3D point is registered to 2D point on the test image. The pose of the test image is optimized by minimizing the re-projection errors. }
        \label{fig_Triangulation_BA}
    \end{figure}
    
The zone label classified for the test image by the Siamese network is further used to retrieve the corresponding pose of the test image. The pose of the median image of that zone from the training data is used to initialize the pose the test image. After the initialization of the pose, the pose is futher refined with several images and poses within the classified zone. Feature points extracted from the stream of images and their corresponding poses are exploited with traditional geometric technique for better estimation of the pose. The two-step pose refinement procedure strictly follows the epipolar geometry to generate the optimal image pose. The overall pose refinement process consists (a) map points triangulation and (b) reprojection based pose refinement. \par

Fig. \ref{fig_Triangulation_BA} shows the procedure of the two-step pose refinement process. 2D corner points on the training images are tracked and matched with a widely used corner points descriptor named Speeded Up Robust Features (SURF) from the stream of selected images within the zone. SURF extractor yields tracked key points and the related properties within the selected image sequence. To further minimize the pose re-projection errors, we utilize the tracked 2D key points in each zone to triangulate the 3D map points. For faster execution, we triangulate the 3D map points in each zone in advance. SURF based feature matching is done for $m$ ($m \geq 3$) images evenly selected from each zone. Further, the labeled poses from $m$ images are regarded as the ground truth poses for the triangulation process. For each registered 2D point $i$ in $j$th image $\mathbf{u}_{ij}\in \mathbb{Z}^2$, we first initialize 3D point by triangulating 2 images and optimize the positions by minimizing the re-projection errors:

\begin{equation}
\begin{split}
\argmin\limits_{\mathbf{v}_1...\mathbf{v}_n} &\sum_{j=1}^m\sum_{i=1}^n \Pi({\mathbf{P}_j},{\mathbf{R}_j},\mathbf{v}_i)-\mathbf{u}_{ij},
\end{split}
\label{Eq_triangulation}
\end{equation}

Where ${\mathbf{P}_j}\in \mathbb{R}^3$ and ${\mathbf{R}_j}\in \mathbb{SO}(3)$ are the position and orientation of the colonoscope at position $j$. $n$ is the number of tracked points. $\mathbf{v}_i\in \mathbb{R}^3$ is the $i$th registered point in 3D space. $\Pi(\cdot)$ project the 3D map points into 2D image space. Further, to cope with outlier, threshold ${\delta}_r$ is applied to filter out 3D map point $\mathbf{v}_i$ with large re-projection error.\par  

With the triangulated 3D map points $\mathbf{v}_i$ and the test image pose initialized from the inference of the Siamese network, we match the map points with the test image, and estimate the corresponding pose of the test image by going through a local Bundle Adjustment (BA). The pose is optimized by minimizing the re-projection errors using Eq. (\ref{Eq_energyfunction}). Where, ${\mathbf{P}}\in \mathbb{R}^3$ and ${\mathbf{R}}\in \mathbb{SO}(3)$  are the target pose of the colonoscope. $\mathbf{v}_i\in \mathbb{R}^3$ is the triangulated 3D map points optimized from Eq. (\ref{Eq_triangulation}). $\mathbf{o}_i\in \mathbb{R}^2$ is the tracked 2D points on the test image. Moreover, we provide an option to handle zone classification errors by filtering out global poses which exceeds the threshold after pose refinement. Using filtering during the two-step pose refinement process provides a robust way to identify the validity of the output and outperform the traditional method in terms of outlier identification. \par

\begin{equation}
\begin{split}
\argmin\limits_{{\mathbf{P}},{\mathbf{R}}} &\sum_{i=1}^m \Pi(\mathbf{v}_i,{\mathbf{P}},{\mathbf{R}})-\mathbf{o}_i.
\end{split}
\label{Eq_energyfunction}
\end{equation}

Eq. (\ref{Eq_triangulation} and \ref{Eq_energyfunction}) are minimized with the Gauss-Newton algorithm. The orientation is represented in the form of Lie group \cite{eade2013lie} and the perturbations are exerted on the left tangent space of the orientations for incremental optimization. After the optimization, thresholds are used for outlier rejection as some zones are textureless and hard to obtain reliable key points for registration. \par

\section{Experimental Results}
\subsection{Dataset}

\begin{figure}[]
        \centering
        \includegraphics[width=0.47\textwidth]{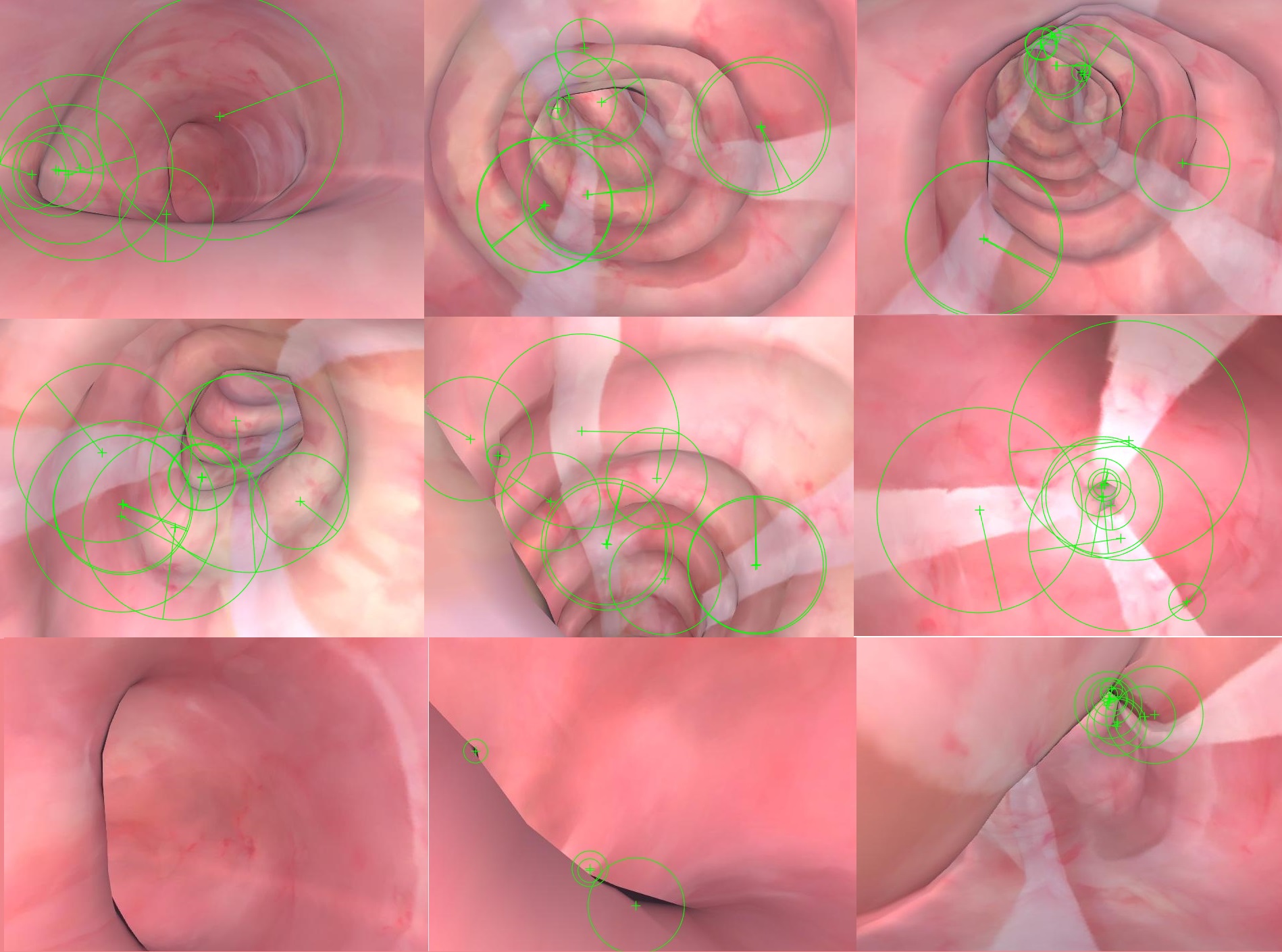}
        \caption{The sample of the generated ex-vivo dataset and the associated extracted corner points. In the upper and middle rows, the number of extracted SURF points exceed 10 while the bottom row only has 0, 4 and 6 SURF points. Note that the SURF extraction is under the same condition: the metric threshold is 200,scale levels set to 6 and octave number is 4.}
        \label{fig:samplesimuimages}
    \end{figure}

To validate the proposed framework, a virtual off-the-shelf phantom\footnote{\url{https://www.turbosquid.com/3d-models/realistic-human-internal-organs-3d-model/1002764}} of a male's digestive system is adopted. A virtual handheld colonoscope is placed inside the colon and is manipulated to go through the colon during which RGB images and the corresponding pose (6 Degree of Freedom (DoF)) is collected. Examples of captured RGB images are shown in Fig.~\ref{fig:samplesimuimages}. We utilize 3D game engine Unity3D\footnote{\url{https://unity.com/}} to generate the sequential RGB images with a pin-hole camera. The virtual phantom is loaded into the Unity3D and the camera is maneuvered within the colon to generate data. The frame rate of the simulated ex-vivo digestive (shown in Fig.~\ref{fig:simulationdateaset_gen}) perspective images is 30 frames per second and the image size is $640 \times 480$. Two separate trajectories are generated\footnote{{The dataset will be open-sourced after the publication.}}, and one is used to train the Siamese network and the other is for testing. The training data consists 2610 images whereas the testing data consists 2603 images ranging from the Anus (the start of the large intestine) to the Appendix (the end of the large intestine). The standard clinical colonoscopy procedure is followed when generating each of these datasets\footnote{Readers are encouraged to refer to the attached video for detailed information regarding the dataset}.\par 

\subsection{Experiment}
The threshold ${\delta}_r$ is set to 10 to filter outlier map points registrations. We extensively test different numbers of zones for our process with Monte Carlo test with each zone tested for ten times. Further, we also test zone division by integrating expert knowledge, where the colon is divided according to the clinical sections shown in Fig. \ref{fig:simulationdateaset_gen} and some sections are divided into sub-classes.\par 

\begin{table*}[]
\caption{Demonstrated are the average error comparisons. The result of classification is presented with the best result. Note that for the classification, we initialize the camera pose by the median pose in the zone. $^*$ means that the approach lose track and the results presented are only partial accuracy. We present our method by integrating expert knowledge for zone classification. The accuracy without expert knowledge (equal zone classification) can be found in Fig. \ref{fig:zones_accuracy_comparison}.}
\label{Table_overall_accuracycompare}
\centering
\begin{tabular}{|c|c|c|c|c|c|c|}
\hline
               & \multicolumn{2}{c|}{Learning based} & \multicolumn{2}{c|}{Geometric based} & \multirow{2}{*}{\begin{tabular}[c]{@{}c@{}}Classification only\\ (Our method)\end{tabular}} & \multirow{2}{*}{\begin{tabular}[c]{@{}c@{}} Pose Refinement \\ (Our method)\end{tabular}} \\ \cline{1-5}
               & PoseNet         & OffsetNet         & ORB-SLAM            & DSO            &                                                                                      &                                                                                        \\ \hline
Position (mm)  & 30.9            & 3.8               & $11.2^*$                & $7.5^*$            & 13.4                                                                                 & \textbf{2.7}                                                                                    \\ \hline
Orientation(°) & 14.76           & 7.11              & $9.36^*$                & $9.1^*$            & 20.83                                                                                & \textbf{6.33}                                                                                   \\ \hline
\end{tabular}
\end{table*}

The Siamese network is trained using the training trajectory consists 2610 images. We randomly choose 128 positive pairs and the same amount of negative pairs in each epoch. The model is trained for $500$ epochs with the learning rate of $0.001$ and batch size of $128$. The training is achieved by minimizing the contrastive loss (Eq. (\ref{Eq_constrastive_loss})), and the weights of the model are optimized using Adaptive movement estimation (Adam) ~\cite{kingma:2014}. Lastly, the model is trained using the Tensorflow libraries~\cite{abadi2016tensorflow} on NIVIDIA titan X GPUs. The weights of the network are initialized with the weights of trained on the Places~\cite{zhou:2014} dataset. Using only the zone classified by the Siamese network and the median of the classified zone as the predicted pose the accuracy achieves $13.4$ mm for position and $20.83$\textdegree~for orientation, whereas the same is $2.7 $ mm and $6.33$\textdegree~respectively.\par

\subsection{Comparison}
\label{sec:comparison}

We compare the proposed framework with DL based regression technique such as PoseNet \cite{kendall2015posenet}, Offsetnet (our implementation) \cite{sganga2018offsetnet} and geometry based methods such as ORB-SLAM \cite{mur2017orb} and DSO \cite{engel2017direct}. 
For fair comparison we would like to address:

\begin{itemize}
    \item For OffsetNet, the training data is treated as the CT dataset with ground truth. OffsetNet is used to calculate the relative position and orientation between the CT image (training image in our case) and RGB image captured by the colonoscope. For the PoseNet, we follow the routine step by feeding images and poses in training.

    \item In the cases of ORB-SLAM and DSO, we align the first pose with ground truth and the scale information from the ground truth is feed into the observation. Moreover, more images are attached at the beginning of the trajectory for better initialization. The images for this are collected by fixing the orientation and moving the camera horizontally.
\end{itemize}

\begin{figure*}
    \centering
    \includegraphics[width=1\textwidth]{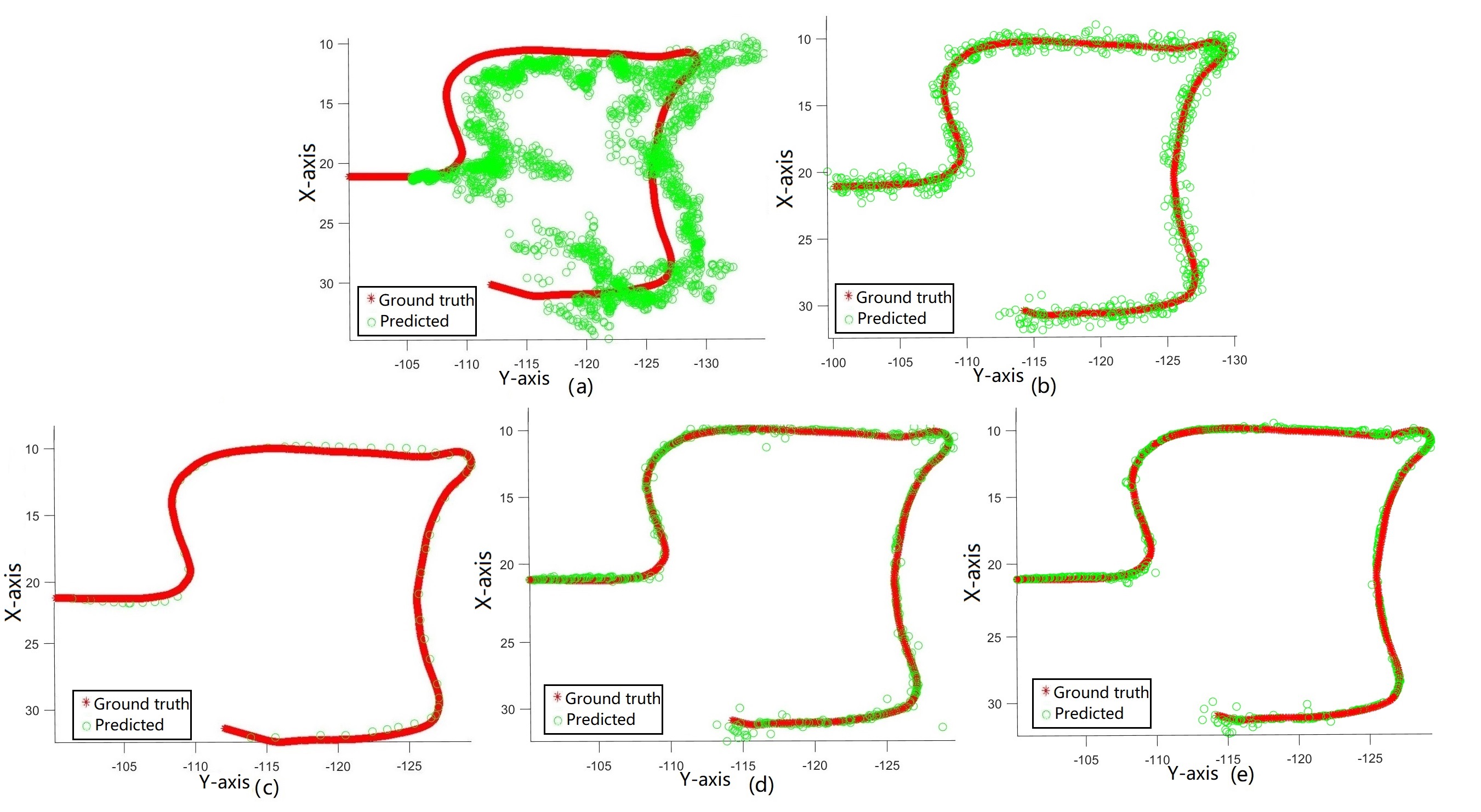}
    \captionof{figure}{We present a method to track monocular images in the GI environment. Note that all figures are unified into X-Y plane because the difference on Z-axis is small and shows fewer differences. (a) is the regressed result from PoseNet. (b) shows the result of Offsetnet. (c) shows the poses initialized to the median of the zone classified by the Siemese network. (d) is the result of colon divided uniformly into 60 zone for classification. (e) is the result considering the expert knowledge (50 zones overall).}
    \label{fig_3Dresults}
\end{figure*}

All the results are presented in Table \ref{Table_overall_accuracycompare} and the corresponding estimated trajectories are shown in Fig. \ref{fig_3Dresults}. Note that the column 'Classification (our method)' only conducts classification and use the median poses in the zone as the predicted pose. Similarly, the prediction in Fig. \ref{fig_3Dresults} (c) is discontinuous and may involves misclassification. Overall, the proposed work with expert knowledge outperforms all previous methods.\par   


\begin{figure*}[]    
    \centering
    \subfloat[]{    
        \begin{minipage}[htpb]{0.4\textwidth}    
            \centering
            \includegraphics[width=1\linewidth]{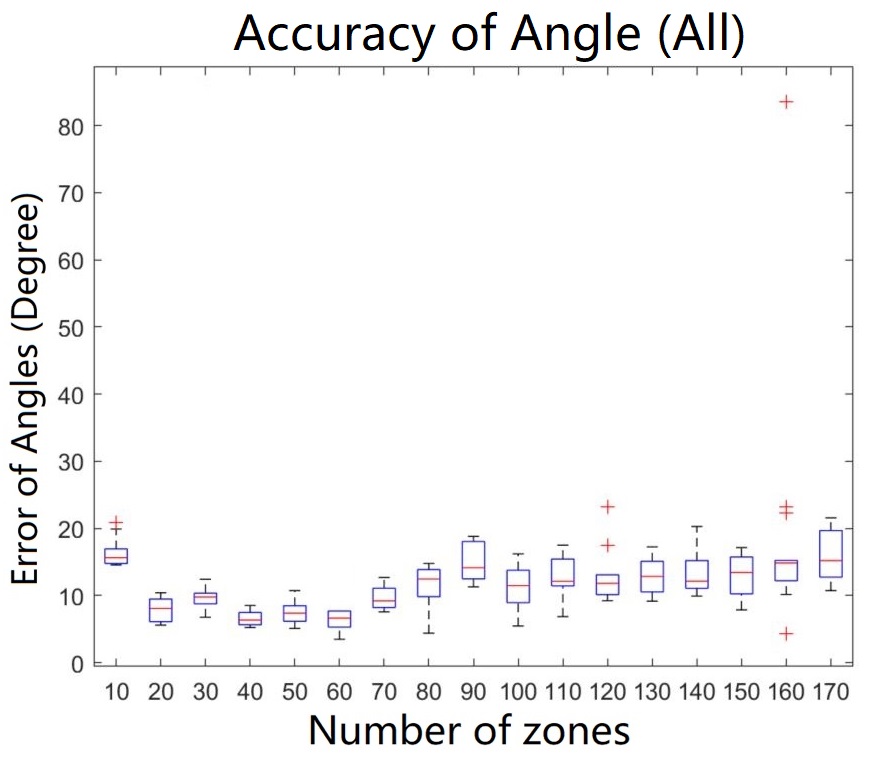}            
        \end{minipage}                
    }\/
    \subfloat[]{
        \begin{minipage}[htpb]{0.39\textwidth}
            \centering
            \includegraphics[width=1\linewidth]{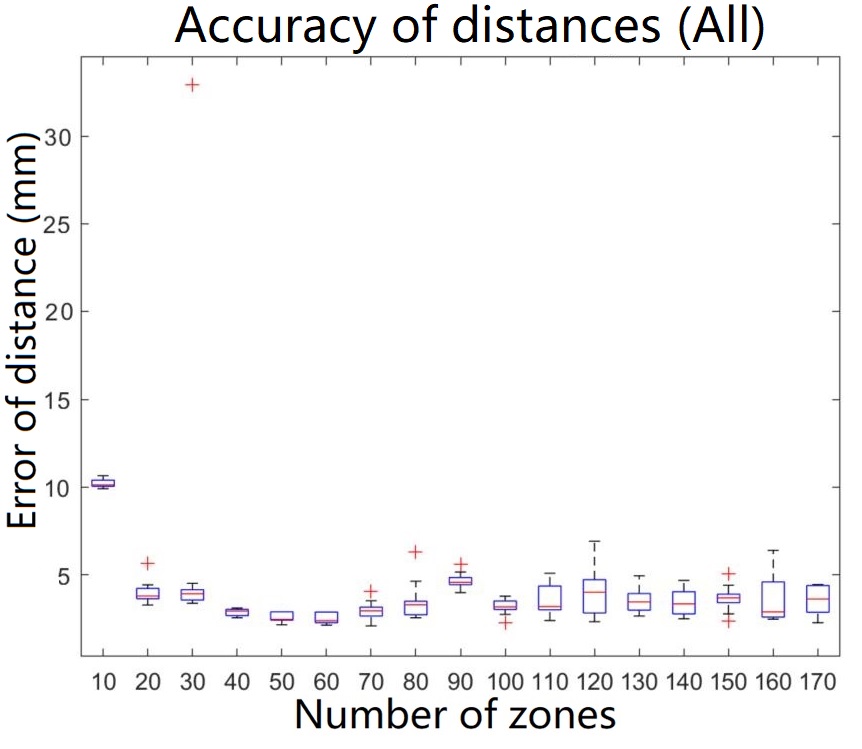}
        \end{minipage}
    }\/

    \caption{Relationship of accuracy and number of zones shown in the Box-and-Whisker chart. Different parameter setting (number of zones) is tested 10 times and validated with the ground truth.}
    \label{fig:zones_accuracy_comparison}
\end{figure*}

\section{Discussion}

The results listed in this paper provide evidential support that using our proposed hybrid method of combining DL with geometric methods provides better estimation accuracy of the pose of a scope in a GI tract. Using our method, the position accuracy improved by 28.94\% when compared with DL methods whereas the same is 64.00\% when compared with the geometric method. Similarly, the orientation accuracy improved by 10.97\% and 30.43\% respectively for DL and geometry based methods. We further elaborate our understanding and impacts of different factors such as zones division of colon, reasons for the failure of DL only and geometry only methods and advantages of the two-step pose refinement.

\subsection{Uniform Zones Vs Expert Knowledge}

To further understand the impact of the number of zones on the overall pose estimation accuracy, we thoroughly perform experiments where the colon is uniformly divided into zones. We perform experiments where the number of zones ranges from 10 to 170 with interval as 10, and each test run 10 times as Monte Carlo experiment. As shown in Fig.~\ref{fig:zones_accuracy_comparison}, our system provides optimal results between 50 and 70 zones. It is driven by two factors. First, the homogeneity of texture within one zone. This significantly affects the performance of the Siamese network, because images with similar texture are naturally more similar. Second, is the size of the zones. Zones with larger coverage limit the co-visible points extracted in the BA process especially for images at front and rear. \par

For completeness, we also compare the performance of our system if the zone division is based on expert knowledge. The off-the-shelf knowledge in large intestine autonomy \cite{gray2009gray} provides information for homogeneity. Texture in different anatomical sections shows different texture \cite{luo2014discriminative} while in the same anatomical zone preserves innate similarity. This knowledge is exploited by generating data that is labeled using expert knowledge. The training trajectory is first classified with authentic large intestine autonomy \cite{gray2009gray} into 13 zones in Fig.~\ref{fig:simulationdateaset_gen}. And some of the larger zones like `Sigmoid flexure', `descending colon', `transverse colon', `ascending colon' are segmented into sub-classes. This strategy enables the best accuracy across all test scenarios shown in Table \ref{Table_overall_accuracycompare}.

\subsection{Comparing With DL Methods}

The result indicates that the accuracy of regression based DL method is inferior to the proposed hybrid method. Overfitting due to the limited amount of training data can be one of the primary causes of failure for DL only method. This is evident as the accumulated position error is $9.2$ mm for training data whereas the same is $30.9$ mm using test data. The limitation of training data in the medical domain is normal due to various privacy issues as against other domains such as autonomous driving. Further, inaccuracy can also be attributed to insufficient overlap caused by small field-of-view of scope. Fig.~\ref{fig_3Dresults} (a-c) shows that various predictions fall outside the colon. Also, previous work \cite{tran2018device,zhang2006image} suggests that enough overlap between consecutive images is required for robust localization. However, for colonoscopy short distance between scope to soft-tissue and relatively fast movement does not guarantee sufficient overlap. Hence, conventional image to pose regression suffers from overfitting and relatively small overlap between images. The proposed Siamese network structure circumvents the problem by relaxing regression with classification. According to Section \ref{section_classification_Siamese}, a training data size with 2610 images and 50 zones will roughly generate 66300 training pairs.\par

The disadvantage of using DL to regress the pose may partially be attributed to the overuse of DL. As \cite{tatarchenko2019single} points out, instead of learning the depth inference from single image (or pose in our scenario), DL only learns the pixel-level classification. Taking DL as a classifier, our work locates the camera pose in the image classification perspective instead of conventional image-pose or image-VO regression. We relax the output from fine-scale pose to coarse-scale pose by considering the essence of DL \cite{tatarchenko2019single}. Thus, the robustness of the process has been greatly promoted and is more explainable. Moreover, the sequential images are divided into several zones; it can be either equally divided or integrated with expert knowledge for interpretability. \par

\subsection{Comparing With Geometric Methods}

There are some disadvantages for geometric feature based methods such as ORB or DSO. First, monocular images do not provide the scale information. Moreover, these methods require sufficient amount of features that can be registered and tracked within images which are challenging especially for textureless soft-tissues. Fig.~\ref{fig:samplesimuimages} shows sample images from our dataset with SURF feature extracted. The SURF features extracted from textureless images shown in the bottom are either limited or located on the viewing edges which are incorrect. Hence, DL methods prevent the losing track issue. \par

\subsection{Two-step Pose Refinement vs Classification Only}

To further validate the two-step pose refinement in our approach, we also compare estimation accuracy by using only the zone classification from the Siamese network. For a fair comparison, we use the median pose of the predicted zone as the pose of the test image. The plot in Fig.~\ref{fig_3Dresults} (c) shows the predicted pose is aligned well with ground truth but in reality the position error especially the orientation error is large as the pose is assigned with the pose of median images within the zone. \par

\begin{figure}[]    
    \centering
    \subfloat[The perturbation impact on the distance.]{    
        \begin{minipage}[htpb]{0.47\textwidth}    
            \centering
            \includegraphics[width=1\linewidth]{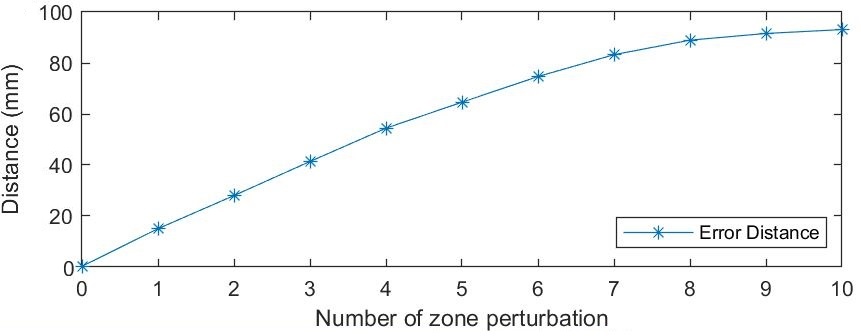}            
        \end{minipage}                
    }\/
    \subfloat[The perturbation impact on the orientation.]{
        \begin{minipage}[htpb]{0.47\textwidth}
            \centering
            \includegraphics[width=1\linewidth]{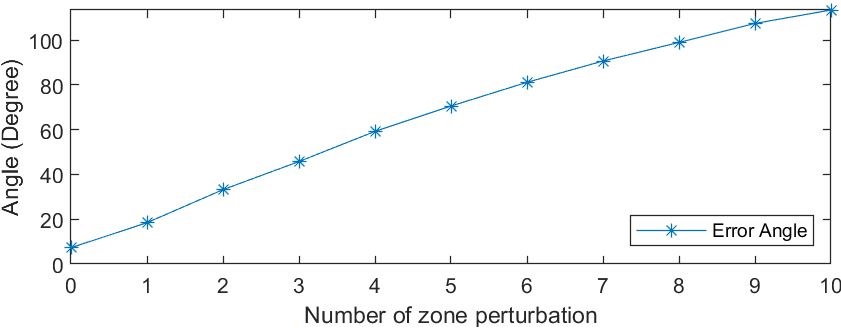}
        \end{minipage}
    }\/
    \caption{The impact of initialization on zone classification. The X-axis shows the arbitrary zone assignment perturbation on the results of classification. The Y-axis presents the refinement errors in distance and orientation.}
    \label{fig:perturbation_on_zones}
\end{figure}

The primary advantage of using the zone classification is that it provides a good initialization for the pose refinement process. 
The two-step refinement accuracy is highly dependent on the quality of initialization as shown in Fig. \ref{fig:zones_accuracy_comparison}. This is primarily due to small parallax angle caused by the special tube like topology in the large intestine. The tracked points located in the center of the image have small parallax and lead to large 3D pose uncertainty. Good initialization significantly contributes to optimal estimation. We conduct a toy test by arbitrarily enforcing wrong classification on our system. We set the perturbation number as $\mathrm{e}$. If the classified zone is indexed in $\mathrm{k}$, the zone index $\mathrm{k}\pm \mathrm{e}$ is assumed as the final classified zone index. Fig. \ref{fig:perturbation_on_zones} shows the impact of this perturbation on final refined pose. Even with $\mathrm{e}=1$ imposed on all predictions, the average error sharply increases from $2.89$ mm to $15.02$ mm and $7.28$\textdegree to $18.46$\textdegree. Thus, the correct initialization of the pose is vital to achieve optimal pose estimates.
 
\section{Conclusions}

In this paper, we propose a hybrid method to combine DL with geometric method to localize colonoscope in the GI tract using monocular images. A Siamese network is used to classify an image to the correct zone for pose initialization. The DL model can be trained with a reasonably small amount of training data. A two-step geometric approach is further used to refine the pose of the test image. Thorough comparisons with other DL and feature based methods further support that our hybrid method provides better pose estimates and is robust to monocular images which possess characteristics such as limited field of view, and/or textureless surface. Lastly, our analysis also suggests that by dividing the colon into zones using expert knowledge provides better performance when compared with equal zone division. 


\bibliographystyle{ieeetr}
\bibliography{endoscope}

\end{document}